\documentclass[11pt,a4paper]{article}
\usepackage[hyperref]{emnlp-ijcnlp-2019}
\usepackage{times}
\usepackage{latexsym}
\usepackage{paralist}
\usepackage{url}

\aclfinalcopy % Uncomment this line for the final submission

\usepackage{adjustbox}
\usepackage{array}
\usepackage{breakurl}

\newcolumntype{R}[2]{%
    >{\adjustbox{angle=#1,lap=\width-(#2)}\bgroup}%
    l%
    <{\egroup}%
}
\newcommand*\rot{\multicolumn{1}{R{90}{1em}}}% no optional argument here, please!

\title{Answers Unite!\\Unsupervised Metrics for Reinforced Summarization Models}

\author{Thomas Scialom$^{\star \ddagger}$, Sylvain Lamprier$^{\ddagger}$, Benjamin Piwowarski$^{\diamond \ddagger}$, Jacopo Staiano$^{\star}$ \\
$^\diamond$ CNRS, France\\
$^\ddagger$ Sorbonne Universit\'e, CNRS, LIP6, F-75005 Paris, France\\
$^\star$ reciTAL, Paris, France \\
  {\tt \{thomas,jacopo\}@recital.ai} \\
  {\tt \{sylvain.lamprier,benjamin.piwowarski\}@lip6.fr} \\}

\date{}

\begin{document}
\maketitle
\begin{abstract}
 Abstractive summarization approaches based on Reinforcement Learning (RL)  have recently been proposed to overcome classical likelihood maximization. % on reference summaries. 
 RL enables to consider complex, possibly non-differentiable, metrics that globally assess the quality and relevance of the generated outputs.
 ROUGE, the most used summarization metric, is known to suffer from bias towards lexical similarity as well as from suboptimal accounting for fluency and readability of the generated abstracts.
 We thus explore and propose alternative evaluation measures: the reported human-evaluation analysis shows that the proposed metrics, based on Question Answering, favorably compares to ROUGE -- with the additional property of not requiring reference summaries.
 Training a RL-based model on these metrics leads to improvements (both in terms of human or automated metrics) over current approaches that use ROUGE as a reward. 
 
\end{abstract}

\section{Introduction}
Summarization systems aim at generating relevant and informative summaries given a variable-length text as input. They can be roughly divided under two main categories, those adopting an \emph{extractive} approach, \emph{i.e.} identifying the most informative pieces from the input text and concatenating them to form the output summary; and those producing \emph{abstractive} summaries, \emph{i.e.} generating an output text whose tokens are not necessarily present in the input text. 

While closer to human summarization, abstractive summarization is a much harder task and the need for faithful evaluation metrics is crucial
to measure and drive the progress of such systems. The %\emph{de-facto} 
standard for evaluation of summarization systems is ROUGE \cite{lin2004rouge}: this metric can be considered as an adaptation of BLEU \cite{papineni2002bleu}, a scoring method for evaluation of machine translation systems; both based on \emph{n}-gram co-occurrences, the latter favors precision while the former emphasizes recall. 

Recent research works \cite{paulus2017deep, pasunuru2018multi, arumae2019guiding} have proposed to use evaluation metrics -- and ROUGE in particular -- to learn the model parameters through Reinforcement Learning (RL) techniques. This makes the choice of a good evaluation metric even more important. Unfortunately, ROUGE is known to incur several problems: in particular, its poor accounting for fluency and readability of the generated abstracts, as well as its bias towards lexical similarity \cite{ng2015better}.
To emphasize the latter point, since ROUGE evaluates a summary against given human references, summarization models incur the risk of being unfairly penalized: a high quality summary might still have very few tokens/\emph{n}-grams in common with the reference it is evaluated against.

In this work, we propose to overcome \emph{n}-gram matching based metrics, such as ROUGE, by developing metrics which are better predictors of the quality of summaries. The contributions of this paper can be summarized as follows: 
\begin{itemize}
    \item Extending recent works \cite{matan2019, chen2018semantic}, we introduce new metrics, based on Question Answering, that do not require human annotations. 
    \item We report a quantitative comparison of various summarization metrics, based on correlations with human assessments. 
    \item We leverage the accuracy of the proposed metrics in several reinforcement learning schemes for summarization, including two unsupervised settings: \emph{in-domain} (raw texts from the target documents) and \emph{out-of-domain} (raw texts from another document collection). 
    \item Besides a quantitative evaluation of the generated summarizes, we %demonstrate
    qualitatively evaluate the performances of the different approaches through human assessment. 
\end{itemize}

Our main results can be summarized as follows:
\begin{enumerate}
    \item We show that fitting human judgments from carefully chosen measures allows one to successfully train a reinforcement learning-based model, improving over the state-of-the-art (in terms of ROUGE and human assessments).
    \item we show that dropping the requirement for human-generated reference summaries, as enabled by the proposed metrics, allows to leverage texts in a self-supervised manner and brings clear benefits in terms of performance.
\end{enumerate}

Section \ref{sec:metrics} introduces the metrics. Section \ref{sec:models} reviews related summarization systems and presents our proposed approaches. Section \ref{sec:experiments} presents our experimental results and discussions. 

\section{Evaluation Metrics} 
\label{sec:metrics}

This section first describes our selection of existing summarization metrics and introduces our proposals. Then, we  quantitatively compare them for abstractive summarization. 
For a comprehensive list of evaluation metrics, we refer the reader to \citet{liu2016not}.

\subsection{\emph{n}-grams-based metrics}

\paragraph{Text-Rank}
Automated summarization started with the development of extractive text summarization models.
Many unsupervised models, that aim at computing a score between a sentence and document(s) were developed -- the score
attempting to reflect whether the sentence should be selected for building a summary \cite{NenkovaAutomaticSummarization2011}.
Such scores can thus be used as a proxy of the summary quality.
We chose Text-Rank \cite{mihalcea2004textrank} -- an extractive non-parametric summarization system inspired by PageRank \cite{page1999pagerank} -- since it is well performing for extractive tasks and could be easily adapted for our needs. The algorithm builds a graph of sentences within a text based on their co-occurrences. Then, it assigns an  importance score for each sentence based on a random walk on the resulting graph. 
The most important elements of the graph are considered as the ones that best describe the text.
As a derivative usage, we propose to consider these importance scores %can be used 
to assess the quality of abstractive summaries in our study. 
This metric is referred to as Text-Rank in the following.

\paragraph{ROUGE} Arguably the most popular metric for summarization at the moment,
it provides a set of measures to compare automatically generated texts against one or more references \cite{lin2004rouge}. In particular, ROUGE-N is based on the count of overlapping \emph{n}-grams, while ROUGE-L accounts for the longest common sub-sequence between the candidate and its corresponding reference(s).

\paragraph{Novelty}
As noted by \citet{see2017get}, abstractive summarization models do not produce novel \emph{n}-grams as often as the reference summaries. 
Thus, to favor the generation of unseen words and produce more abstractive summaries, \citet{kryscinski2018improving} integrated \emph{novelty} as a reward for reinforcement learning. It is defined as the fraction of unique \emph{n}-grams in the summary that are novel, normalized by the length ratio of the generated and reference summaries.

\subsection{Beyond n-grams}

\subsubsection{Language Modeling}
We investigate the use of language models as an evaluation metric. \cite{shafieibavani-etal-2018-summarization} proposed to exploit word embeddings to train a model able to rate the generated summaries. 
Following neural language models (LM), we propose to consider the perplexity of the generated summary according to 
the BERT LM \cite{devlin2018bert}, which demonstrated state of the art results in many NLP tasks. For our experiments, we used the publicly available pre-trained English ``base'' model.  

\subsubsection{Question-Answering based Metrics}

Question-Answering is related to summarization: the first work in this direction \cite{wu2002towards} introduced the notion of Answer-Focused Summarization, where answers to relevant questions on the source text are used to build the corresponding summary.
Based on the intuition that a good-quality summary should provide the answers to the most relevant questions on a given input text, several works have proposed to adapt Question Answering (QA) for summary quality evaluation.

In that vein, \cite{pasunuru2018multi}  proposed to measure if answers contain the most \emph{salient} tokens. 
Closer to our work, \cite{matan2019} proposed \emph{APES}, a novel metric for evaluating summarization, based on the hypothesis that the quality of a generated summary is linked to the number of questions (from a set of relevant ones) that can be answered by reading it. %
In their proposed setup, two components are thus needed: (a) a set of relevant questions for each source document; and (b) a QA system.  For each summary to assess, questions are successively generated from a reference summary, by masking each of the named entities present in this reference,  %the input text 
following the methodology described in \cite{hermann2015teaching}. This results in as many triplets $(input, question, answer)$ as named entities present in the reference summary, where \emph{input} denotes the summary to assess, \emph{question} refers to the sentence containing the masked entity and \emph{answer} refers to this masked entity to retrieve. 
Thus, for each summary to assess, metrics can be derived from the ability of the QA system to retrieve the correct answers from each of the associated triplets.

\paragraph{F1 score}

For each triplet, an F1 score is computed according to the responses retrieved by the considered QA system. This score, commonly used for QA evaluation \cite{rajpurkar2016squad}, measures the average overlap between predictions and ground truth answers. For each summary to assess, the is the average of the F1 score computed over each triplet. In the following, we denote this metric as \emph{QA\textsubscript{$fscore$}(sup)}.

\paragraph{QA confidence}  

Complementary to the F1 score, we propose to also consider the confidence of the QA system for its retrieved answer. This corresponds, for each triplet, to the probability of the true answer according to the QA model.
Confidence scores are averaged for each summary to assess over its associated triplets. In the following, we denote this metric as \emph{QA\textsubscript{$conf$}(sup)}.

Besides considering the simple presence of the expected answers in the generated summary, QA-based metrics also  account to some extent for readability. They indeed require that the considered  QA system, trained on natural language, be able to find the answer in the input to assess, despite the variability of the generated texts. 

\paragraph{Extension to the unsupervised setting} 
While being a useful complement to ROUGE, the two QA-based metrics described above still need human-generated summaries. 
In this paper, we investigate and propose extending the previously described QA-based approach in an unsupervised setting. 
 
With this aim, we extended the above metrics at \emph{the document level} (\emph{i.e.}, questions and answers are generated from the source article text rather than from the reference summary), dispensing of the need for  human-generated reference summaries. 
Thus, in line with the \emph{APES} approach described above, we propose two unsupervised QA-based metrics, to which we refer to as \emph{QA\textsubscript{$fscore$}(unsup)} and \emph{QA\textsubscript{$conf$}(unsup)}. Accounting for both quality and informativeness of a generated summary, those metrics have the appealing property of not requiring reference summaries.

\subsection{Quantitative Analysis}
We exploit human judgments obtained for 3 types of automatically generated summaries by \citet{paulus2017deep} on 100 samples of the CNN/Daily Mail summarization dataset (see detail in section \ref{sec:data}), in terms of readability (how well written the summary is) and relevance (how well does the summary capture the important parts of the article). 
The summaries are generated by the three different systems proposed in the original work.
Those samples have been scored, via Amazon Mechanical Turk, for Readability and Relevance (scores from 1 to 10 for both metrics). 

\begin{table}[t!]
\begin{center}
\begin{tabular}{|l|r|r|}
\cline{2-3}
    \multicolumn{1}{c|}{}
                        & \textbf{Readability} & \textbf{Relevance}\\
\hline
\textbf{Readability}     & 1.0         & 0.77 **\\
\textbf{Relevance}       & 0.77 **     & 1.0\\
\hline
\hline
ROUGE-1 (sup)            & 0.14 *      & 0.18 **\\
ROUGE-2 (sup)            & 0.12 *      & 0.18 **\\
ROUGE-L (sup)            & 0.13 *      & 0.18 **\\
Text-Rank (unsup)         & 0.14 *      & 0.13 **\\
Novelty (unsup)          & -0.13 *     & -0.1 * \\
\hline
\hline
Bert LM (unsup)                  & 0.21 **     & 0.08 *\\
QA\textsubscript{$fscore$} (sup)   & 0.14 *      & 0.19 **\\
QA\textsubscript{$conf$} (sup)     & 0.19 **     & 0.23 **\\
QA\textsubscript{$fscore$} (unsup) & 0.08        & 0.2 ** \\
QA\textsubscript{$conf$} (unsup)   & 0.33 **     & 0.31 **\\
\hline
\end{tabular}
\end{center}
\caption{Spearman's $\rho$ for the different metrics w.r.t. Readability and Relevance (*: $p<.05$, **: $p<.005$).}
    \label{tab:correlation_metrics_human}
\end{table}

In Table~\ref{tab:correlation_metrics_human}, we report  Spearman's rank correlations on this data, where we compare  summaries rankings obtained according to the assessed metrics. % with rankings obtained following both human assessments. 
Scores render the ability of the various metrics to reproduce human preferences (in terms of readability and relevance). 
First, we observe that readability and relevance are naturally intertwined: intuitively, an unreadable summary will bear very little information, one of the facts that explains the high correlation between \emph{readability} and \emph{relevance}. 

From this correlation analysis against human judgments, we observe that, as expected, the Language Model metric captures \emph{readability} better than ROUGE, while falling short on \emph{relevance}. 

On the other hand, the results obtained using the proposed QA-based metrics indicate their potential benefits especially under the \emph{unsupervised} setting, with \emph{QA\textsubscript{$conf$}} and \emph{QA\textsubscript{$fscore$}} capturing \emph{readability} and \emph{relevance} better than all the others reported metrics, including ROUGE.
We thus conclude that the proposed metrics, which favorably correlate with \emph{readability} and \emph{relevance} under human evaluation, are worth of a deeper experimental investigation: in the following  sections we provide a thorough evaluation of their contributions as Reinforcement Learning rewards signals.

\subsection{Learned Metric}
\label{sec:learnedmetric}
Finally, we also leverage the qualitative data obtained by \citet{paulus2017deep} -- which compounds to 50 samples evaluated by annotators in terms of \emph{readability} and \emph{relevance} -- to \emph{learn} an aggregate metric for evaluation.
We use a Ridge regression (with a regularization $\lambda=1$) to learn to predict the geometric mean of readability and relevance from the metrics defined above. The geometric means was chosen since we want the generated summary to be both readable and relevant.

We randomly sampled 50\% of the data to fit the linear model with various subsets of our base metrics. Then, we measured the correlation w.r.t. the expected geometric mean on the remaining 50\% data. We performed this procedure 1000 times. 
Our experiments show that the best performing set of metrics consists of ROUGE-L in conjunction with QA\textsubscript{$conf$} and QA\textsubscript{$fscore$}, both computed at an article-level, and hence unsupervised. 

This learned metric is thus defined as (with \emph{unsup} versions of QA-based scores):
\begin{equation}
\label{eq:learned}
\alpha ROUGE_L + \beta QA_{conf} + \delta QA_{fscore}
\end{equation}

\noindent with $\alpha=0.8576$, $\beta=2.274$ and $\delta=0.6413$.
We leverage this learned metric in our RL-based summarization model, as described below.     

\subsection{Implementation details}
As QA system we use the BERT ``base" pre-trained model \cite{devlin2018bert}, finetuned on the SQuAD dataset \cite{rajpurkar2016squad} using the recommended parameters for the task\footnote{\url{https://github.com/huggingface/pytorch-transformers}.}. This differs from the approach adopted by \cite{matan2019} who trained their QA model on CNN-DM (the same data used for the summarization task).

\section{Summarization Models} \label{sec:models}

Abstractive summarization systems were originally designed as a post-processing of an extractive system -- by compressing sentences \cite{NenkovaAutomaticSummarization2011}. A lot of activity takes place nowadays in designing neural networks sequence to sequence architectures \cite{sutskever2014sequence}, which allow to consider the problem as a whole rather than a two-step process. Such models reached state-of-the-art performance. To tackle the summarization, which deals with a long text and possibly out-of-vocabulary tokens, \citet{see2017get} proposed to leverage an attention over the input \cite{bahdanau2014neural}, as well as a copy mechanism \cite{vinyals2015pointer}. 

One problem of sequence-to-sequence models is that they tend to repeat text in the output. To deal with this problem,
\cite{see2017get} use a \emph{coverage mechanism}, and \citet{paulus2017deep} introduced \emph{Intra-Decoder Attention} with the same goal of avoiding duplicate information within the output sequences. 

More recently, the model proposed by \citet{see2017get} was further extended \cite{gehrmann2018bottom}, with the addition of an attention mask during inference: a pre-trained sequence tagger trained to select which input tokens should be copied and used to filter the copy mechanism. Such a filter, called \emph{Bottom-Up Copy Attention}, was shown to help prevent copying from the source text sequences that are too long, hence resulting in more abstractive summaries. On the CNN/Daily Mail dataset, \cite{gehrmann2018bottom} found this two-step process to yield significant improvements in terms of ROUGE -- resulting in the current state-of-the-art system. We base our experiments on this model.

The differentiable loss function commonly used for training summarization models, negative log-likelihood, has several known limitations.
Among those, exposure bias and failure to cope with the large number of potentially valid summaries. 

To overcome this, approaches based on reinforcement learning have recently been proposed, allowing the models to learn via reward signals.
\citet{ranzato2015sequence} used the REINFORCE algorithm \cite{williams1992simple} to train RNNs for several generation tasks, showing improvements over previous supervised approaches. 
\citet{narayan-etal-2018-ranking} used such an approach in an extractive summarization setting, learning to select the most relevant sentences within the input text in order to construct its summary. \cite{paulus2017deep} combined supervised and reinforcement learning, demonstrating improvements over competing approaches both in terms of ROUGE and on human evaluation. 
However, the main limit of these works is that they rely on standard summarization metrics which are known to be biased.

Finally, closer to our work, \citet{arumae2019guiding} proposed to use question-answering rewards to learn an \emph{extractive} summarization model in a reinforcement learning setup. Compared to what we propose, their system is extractive, and relies on hand-written summaries.

\subsection{Mixed Training Objectives}

In our experiments, we follow the reinforcement learning scheme 
described below. The main difference with previous works is our reward function, which was based on our study of metrics (section \ref{sec:metrics}).
We consider a mixed loss $L_{ml+rl}$ combining supervised and reinforcement learning schemes:
\begin{equation}
L_{ml+rl} = \gamma L_{rl} + (1 - \gamma) L_{ml}
\label{mlrl}
\end{equation}
where we define the reinforcement loss $L_{rl}$ and the maximum likelihood $L_{ml}$ in the following paragraphs.

\paragraph{Maximum Likelihood}
Under a supervised training setup, the teacher forcing algorithm~\cite{williams1989learning} can be applied, and corresponds to maximizing the likelihood (ML) or 
equivalently to minimizing the negative log likelihood (NLL) loss defined as:
\begin{equation}
L_{ml} = -\sum^{m}_{t=0}{\log p(y^{*}_t | y^{*}_0, ..., y^{*}_{t-1}, X)}    
\end{equation}
where $X=[x_1,...,x_n]$ is the input text of $n$ tokens and $Y^*=[y_1^*,...,y_m^*]$ is the corresponding reference summary of $m$ tokens. 

\paragraph{Policy Learning}
Several RL-based summarization \cite{kryscinski2018improving, li2018actor, pasunuru2018multi, paulus2017deep} apply the self-critical policy gradient training algorithm \cite{rennie2017self}. Following \cite{paulus2017deep} we use REINFORCE algorithm, using as the baseline a greedy decoding algorithm according to the conditional distribution $p( y | X )$, giving rise to a sequence $\widehat{Y}$. The model is sampled using its Markov property, that is, one token at a time, giving rise to the sequence $Y^s$.

Following the standard RL actor-critic scheme, with $r(Y)$ the reward function for an output sequence Y, the loss to be \emph{minimized} is then defined as:
\begin{equation}
L_{rl} =  (r(\widehat{Y}) - r(Y^s)) \sum^{m}_{t=0}{\log p(y^{s}_t | y^{s}_0, ..., y^{s}_{t-1}, X)}
\end{equation}

As ROUGE is the most widely used evaluation metric, \citet{paulus2017deep} used ROUGE-L as the reward $r$ for the $L_{rl}$ function and tested the following three different setups: 

\begin{itemize}
    \item{\emph{ML}: the model trained with $L_{ml}$ ($\gamma=0$)};
    \item{\emph{RL}: the model trained with $L_{rl}$ ($\gamma=1$)};
    \item{\emph{ML+RL}: the model trained with $L_{ml+rl}$ ($\gamma=0.9984$)}.
\end{itemize}

The human evaluation conducted on the three models shows that \emph{RL} performs worse than \emph{ML}, and \emph{ML+RL} performs best for both \emph{readability} and \emph{relevance}. The authors also conclude that ``despite their common use for evaluation, ROUGE scores have their shortcomings and should not be the only metric to optimize on summarization model for long sequences", which is translated in the very high optimal $\gamma$. We show that using a more sensible metric to optimize leads to a better model, and to a lower $\gamma$.

\section{Experiments} 
\label{sec:experiments}

In our experiments, we evaluate the effect of substituting the ROUGE reward in the reinforcement-learning model of \cite{paulus2017deep} by our proposed metric (section \ref{sec:metrics}). We, moreover, study the effect of using metrics that do not necessitate human-generated summaries.

\subsection{Data Used} \label{sec:data}
Task-specific corpora for building and evaluating summarization models associate a human-generated reference summary with each text provided. 
We resort to the CNN/Daily Mail (CNN-DM) dataset \cite{hermann2015teaching,nallapati2016abstractive} for our experiments. It includes 287,113 article/summary pairs for training,  13,368 for validation, and 11,490 for testing. The summary corresponding to each article consists of several bullet points displayed on the respective news outlet webpage. In average, summaries contain 66 tokens ($\sigma=26$) and 4.9 bullet points. Consistently with \citet{see2017get} and \citet{gehrmann2018bottom}, we use the non-anonymized version of the dataset, the same training/validation splits, and perform truncation of source documents and summaries to 400 and 100 tokens, respectively.

To assess the possible benefits of reinforcing over the proposed QG-based metric, which does not require human-generated reference summaries, we employ TL;DR\footnote{\url{https://tldr.webis.de}}, a large-scale dataset for automatic summarization built on social media data, compounding to 4 Million training pairs \cite{volske2017tl}. Both CNN-DM and TL;DR datasets are in English.

\subsection{Models}

For all our experiments, we build on top of the publicly available OpenNMT implementation\footnote{\url{http://opennmt.net/OpenNMT-py/Summarization.html}}, consistently with \citet{gehrmann2018bottom} to which we refer to as a baseline. 
The encoder is composed of a one-layer bi-LSTM with 512 hidden states, and 512 hidden states for the one-layer decoder. The embedding size is set at 128. The model is trained with Adagrad, with an initial learning rate of 0.15, and an initial accumulator value of 0.1. We continue training until convergence; when the validation perplexity does not decrease after an epoch, the learning rate is halved. We use gradient-clipping with a maximum norm of 2. 

\citet{gehrmann2018bottom} showed that increasing the number of hidden states leads to slight improvements in performance, at the cost of increased training time; thus, as reinforcement learning is computationally expensive, we build on top of the smallest model -- nonetheless, we include the largest model by \citet{gehrmann2018bottom} in our discussion of results.

All the experimented reinforcement approaches use the mixed training objectives defined in 
equation \ref{mlrl}, with the ML part corresponding to the previously described baseline model pretrained on the CNN-DM dataset.
Compared models differ on the considered reward signals. They also differ on their use of additional unsupervised data, either \emph{In-Domain} or \emph{Out-of-Domain}, as discussed below. 

\subsubsection{Reward Signals}
The three reward signals used throughout our experiments, are detailed below: 
\begin{enumerate}
    \item \textbf{ROUGE}: We use only ROUGE-L as reward signal within the baseline architecture, consistently with \citet{paulus2017deep};
    \item \textbf{QA\textsubscript{$learned$}}: Conversely, we compute the reward by applying the learned coefficients to the three components of the learned metric, as obtained in Section~\ref{sec:learnedmetric}.
    \item \textbf{QA\textsubscript{$equally$}}: We apply the mixed training objective function, using as a reward the three metric components of the learned metric (ROUGE-L, QA\textsubscript{$conf$}, and QA\textsubscript{$fscore$}) equally weighted: this corresponds to setting a value of 1 for $\alpha$, $\beta$ and $\delta$ in Eq.~\ref{eq:learned}. This allows to see to which extent learning is sensitive to fitting human assessments.
\end{enumerate}

For (2) and (3), we set $\gamma$  (Eq. \ref{mlrl}) to 0.5\footnote{We have run experiments with $\gamma=0.5$, and $\gamma=0.9984$ as \citet{paulus2017deep}; we report here the best performance which was obtained with the former.}. This shows that, compared to \cite{paulus2017deep}, we do not need to use NLL to avoid the model from generating unreadable summaries.

\subsubsection{In-Domain vs Out-of-Domain}
\label{sec:inoutdomain}
Finally, we experiment with the proposed QA\textsubscript{$conf$} and QA\textsubscript{$fscore$} metrics in an unsupervised fashion, as they can be computed at article level -- \emph{i.e.} without accessing the reference human-generated summaries.
We investigate the potential benefits of using this approach both \emph{in-domain} and \emph{out-of-domain}: for the former, we resort to the test set of the CNN-Daily Mail (CNN-DM) dataset; for the latter, we leverage the TL;DR corpus. 

As the CNN-Daily Mail is built from mainstream news articles, and the TL;DR data comes from social media sources, we consider the latter as out-of-domain in comparison. From the latter, which includes circa 4 million samples, we randomly draw sample subsets of size comparable  with CNN-DM for training, validation and testing splits.

Due to computational costs, we restrict these experiments to the model trained under reinforcement using the QA\textsubscript{$learned$} metric. 
Under this setup, the model has access at training time both to:
\begin{itemize}
    \item \emph{supervised} samples for which a reference summary is given (and thus all metrics, including ROUGE and NLL, can be computed as a training objective), coming from the training set of CNN-Daily Mail corpus ;
    \item \emph{unsupervised} samples, for which no reference is available, thus allowing to only compute QA\textsubscript{$conf$}(unsup) and QA\textsubscript{$fscore$}(unsup). Three unsupervised settings are considered in the following:
    
    \emph{TL;DR}, corresponding to the \emph{out-of-domain} setting where we use articles from the \emph{TL;DR} dataset;
    
    \emph{CNN-DM (VAL)}, corresponding to an \emph{in-domain} setting where we use texts from the validation set from the CNN/Daily Mail dataset;
    
    and, \emph{CNN-DM (TEST)} for an \emph{in-domain} setting  where we use the articles from the test set (thus containing texts used for evaluation purposes).  
\end{itemize}

While all the data is from the CNN-DM train dataset in the \emph{supervised} setups, for the \emph{unsupervised} setups, we set the proportion of unsupervised data to 50\% (either CNN-DM VAL,  CNN-DM TEST for \emph{in-domain} or TL;DR for \emph{out-of-domain} data).
Thus, for 50\% of the data, the model has access only to the QA\textsubscript{$conf$} and QA\textsubscript{$fscore$} reward signals, since the ROUGE-L reward can only be computed on \emph{supervised} batches. 

Therefore, for all the unsupervised setups, in order to keep consistency in the reward signal throughout the training, we multiply by a factor of 2 the weight associated with ROUGE-L when this reward is computable, and set it to 0 otherwise.

\subsection{Results}

\begin{table*}[t!]
\begin{center}
\begin{tabular}{|l|r|r|r||r|r|}
\cline{2-6}
    \multicolumn{1}{c|}{}
                        & \textbf{R-1} & \textbf{R-2} & \textbf{R-L} & \textbf{QA\textsubscript{$fscore$}} & \textbf{QA\textsubscript{$conf$}} \\

\hline
\citet{see2017get}              & 39.53 & 17.28 & 36.38 &-&-\\
\citet{gehrmann2018bottom}      & 41.22 & \textbf{18.68} & 38.34 &-&-\\
\hline
\hline
ML+RL \citet{paulus2017deep}    & 39.87 & 15.82 & 36.90 &-&-\\
RL \citet{paulus2017deep}       & 41.16 & 15.75 & 39.08 &-&-\\
\citet{pasunuru2018multi}       & 40.43 & 18.00 & 37.10 &-&-\\
\citet{chen2018fast}            & 40.88 & 17.80 & 38.54 &-&-\\
\hline
\hline
baseline                                                     & 42.24 & 17.78 & 37.44 & 14.91 & 40.12\\
~+ ROUGE                                             & \textbf{45.62} & 16.30 & \textbf{41.60} & 13.64 & 37.90\\
~+ QA\textsubscript{$equally$}                   & 43.36 & 18.06 & 38.33 & 16.06 & 41.01\\
~+ QA\textsubscript{$learned$}                & 42.71 & 17.81 & 37.94 & 15.19 & 41.39\\
~+ QA\textsubscript{$learned$} + TL;DR         & 42.75 & 17.57 & 37.88 & 15.75 & 41.54\\
~+ QA\textsubscript{$learned$} + CNN-DM (VAL)          & 43.00 & 17.66 & 38.23 & 16.16 & 41.75\\
~+ QA\textsubscript{$learned$} + CNN-DM (TEST)        & 42.74 & 17.25 & 37.96 & \textbf{16.17} & \textbf{42.14}\\
\hline
\end{tabular}
\end{center}
\caption{Comparison with previous works. On top, we report the results obtained by \citet{gehrmann2018bottom} using their largest architecture, as well as those by \citet{see2017get}. Next, we report results recently obtained by reinforcement learning approaches. Finally, we indicate the scores obtained by our baseline -- the ``small" model by \citet{gehrmann2018bottom} -- and the six reinforced models we build on top of it.}
\label{tab:big_table}
\end{table*}

In Table~\ref{tab:big_table}, we report the results obtained from our experiments in comparison with previously proposed approaches.
We observe that, unsurprisingly, reinforcing on ROUGE-L allows to obtain significant improvements over the state-of-the-art, in terms of ROUGE but at the cost of lower QA-based metrics. Conversely, reinforcing on the proposed metric improves consistently all its components (ROUGE-L, QA\textsubscript{$conf$} and QA\textsubscript{$fscore$}). 

However, increasing the reward does not necessarily correlate with better summaries. The human inspection as reported by \cite{paulus2017deep} shows that the generated summaries reinforced on ROUGE-L are consistently on the low end in terms of readability and relevance. 

A closer inspection of the generated summaries revealed that the sequences generated by this model seem to qualitatively degrade as the number of produced tokens grows: they often start with a reasonable sub-sequence, but quickly diverge towards meaningless outputs. 
This can be explained by the aforementioned drawbacks of ROUGE, which are likely amplified when used both as evaluation and reward: the system might be optimizing for ROUGE, at the price of losing the information captured with the NLL loss by its language model.

We hence conducted a human evaluation for the different setups, reported in Table~\ref{tab:human_evaluation}, assessing their outputs for \emph{readability} and \emph{relevance} in line with \citet{paulus2017deep}. We randomly sampled 50 articles from the CNN-DM test set; since the learned metric used in our experiments is derived from the subset manually evaluated in \citet{paulus2017deep} we ensured that there was no overlap with it. For each of those 50 articles, three English speakers evaluated the summaries generated by the 7 different systems reported in Table~\ref{tab:big_table}.

\begin{table*}[t!]
\begin{center}
\begin{tabular}{|l|r|r|}
\cline{2-3}
    \multicolumn{1}{c|}{}
                        & \textbf{Readability} & \textbf{Relevance} \\
\hline
human reference                                              & 7.27* & 7.4** \\
\hline
baseline                                                     & 7.07 & 5.82 \\
~+ ROUGE                                             & 2.14** & 5.48** \\
~+ QA\textsubscript{$equally$}                   & 5.94** & 6.34** \\
~+ QA\textsubscript{$learned$}                & 6.96 & 6.21** \\
~+ QA\textsubscript{$learned$} + TL;DR         & 6.60* & 6.26** \\
~+ QA\textsubscript{$learned$} + CNN-DM (VAL)          & 6.40* & 6.75** \\
~+ QA\textsubscript{$learned$} + CNN-DM (TEST)        & 6.89 & 6.80** \\

\hline
\end{tabular}
\caption{Human assessment: two-tailed t-test results are reported for each model compared to the baseline ($*: p < .01$, $**: p < .001$).}
\label{tab:human_evaluation}
\begin{tabular}{ll|l|l|l|l|l|l|l}
    & \rot{human reference} & \rot{baseline} & \rot{+ ROUGE} & \rot{+ QA\textsubscript{$equally$} } & \rot{+ QA\textsubscript{$learned$}} & \rot{+ QA\textsubscript{$learned$} + TL;DR} & \rot{+ QA\textsubscript{$learned$} + CNN-DM (VAL)} & \rot{+ QA\textsubscript{$learned$} + CNN-DM (TEST)} \\
\hline
baseline & * / ** & - &  &  &  &  &  \\
~+ ROUGE & ** / ** & ** / ** & - &  &  &  & \\
~+ QA\textsubscript{$equally$} & ** / ** & ** / ** & ** / ** & - &  &  &  \\
~+ QA\textsubscript{$learned$} & ** / ** &  - / ** & ** / ** & ** / - & - &  & \\
~+ QA\textsubscript{$learned$} + TL;DR & ** / ** & * / ** & ** / ** & ** / - & * / - & - & \\
~+ QA\textsubscript{$learned$} + CNN-DM (VAL)  &** / ** & * / ** & ** / ** & ** / * & ** / ** &  - / * & - & \\
~+ QA\textsubscript{$learned$} + CNN-DM (TEST)        & ** / ** &  - / ** & ** / ** & ** / * & - / ** &  - / * & * / - & -\\

\hline
\end{tabular}
\end{center}
\caption{Human assessment: two-tailed t-test results are reported for each model pair for Readability / Relevance ($*: p < .01$, $**: p < .001$).}
\label{tab:significance}
\end{table*}

We observe that reinforcing using the proposed metric -- which includes QA based metrics, leads to comparable performance in terms of ROUGE w.r.t. state-of-the-art approaches, while clear benefits emerge from the results of the human evaluation: a significant improvement in terms of relevance, particularly when leveraging \emph{in-domain} data in an \emph{unsupervised} setup.
Not surprisingly, we observe an improvement for our model when reinforced through the learned metric compared to the one equally weighted. The slightly lower relevance scores observed for the QA\textsubscript{$learned$} w.r.t. QA\textsubscript{$equally$} are consistent with the lower ROUGE-L and QA\textsubscript{$fscore$} reported in Table~\ref{tab:big_table}. This is explained by the lower coefficients for ROUGE-L and QA\textsubscript{$fscore$} (see \ref{sec:learnedmetric}), and the relatively stronger correlation of those two metrics with \emph{relevance} (see Table~\ref{tab:correlation_metrics_human}).

Consistently with the figures reported in Table~\ref{tab:big_table}, the human evaluation results -- reported in Tables~\ref{tab:human_evaluation} and~\ref{tab:significance} -- confirm the progressive improvements of our different proposed models when using unsupervised data closer to the test set documents: 
\begin{itemize}
    \item adding \emph{unsupervised} data from the out-of-domain TL;DR brings a slight improvement using QA\textsubscript{$learned$};
    \item when it comes to the same domain (\emph{i.e.} CNN-DM validation) the improvements increase;
    \item finally, when unsupervised samples come from the same set as those used for testing, we observe even better results.
\end{itemize}
These results show that using the proposed QA-based metrics, that do not depend on reference summaries, allows to leverage raw text data; and, that fine-tuning (without supervision) on the documents to be summarized is beneficial.

To elaborate further, we notice that applying the learned coefficients for~\ref{eq:learned} to the results obtained by models reinforced on QA\textsubscript{learned} and QA\textsubscript{equally}, see Table~\ref{tab:big_table}, we obtain very similar scores (namely, 136.43 for QA\textsubscript{equally} and 136.4 for QA\textsubscript{learned}).
However, the qualitative analysis reported in Tables~\ref{tab:human_evaluation} and~\ref{tab:significance} shows that while they perform similarly in terms of \emph{relevance}, a significantly lower score for \emph{readability} is obtained using QA\textsubscript{equally}. This can be explained by the stronger weight of ROUGE\_L for this setup, a fact which might lead to a degradation of the quality of the output consistently with the observations reported in \cite{paulus2017deep} as well as in our ROUGE experiment. 

Another observation from Tables~\ref{tab:human_evaluation} and~\ref{tab:significance} is that while QA\textsubscript{learned} performs significantly better in term of \emph{readability} than QA\textsubscript{learned} + CNN-DM (VAL), the opposite holds for \emph{relevance}. This could be explained by the setup difference during training: as detailed in section \ref{sec:inoutdomain}, for unsupervised setups (\emph{i.e.} QA\textsubscript{learned} + CNN-DM (VAL)) only the QA-based metrics are computed for the portion of data for which no reference is available. 
While testing (TEST) and validation (VAL) splits come the same dataset (CNN-DM), we observe that using the samples from TEST in an unsupervised fashion allows for maintaining comparably high \emph{relevance} compared to QA\textsubscript{learned} + CNN-DM (VAL), while also obtaining similar \emph{readability} to QA\textsubscript{learned}. This shows the possible benefits that can be obtained by exposing the model to the evaluation data in unsupervised setups.
To further study our unsupervised metrics, we performed additional experiments on the TL;DR corpus. 
We observed more than one absolute point of improvement w.r.t CNN-DM TEST in terms of ROUGE-L, QA\textsubscript{$fscore$} (unsup) and QA\textsubscript{$conf$} (unsup). 

This indicates that the proposed unsupervised metrics allow the  model to better transfer to new domains such as TL;DR.
These results pave the way for leveraging large numbers of texts, in a self-supervised manner, to train automatic summarization models. %

\section{Conclusions}
We have presented the analysis of novel QA-based metrics\footnote{A python package will be made available at \url{https://www.github.com/recitalAI/summa-qa}.}, and have shown promising results when using them as a reward in a RL setup. Crucially, those metrics do not require a human reference, as they can be computed from the text to be summarized. 

From our experiments this proves particularly beneficial, allowing to leverage both \emph{in-domain} and \emph{out-of-domain} unlabeled data.

The promising results obtained indicate a path towards partially self-supervised training of summarization models, and suggest that progress in automated question generation can bring benefits for automatic summarization.

Finally, to our knowledge, this paper is the first to compare two architectures with the same reinforcement setup on the same data: the one proposed by \citet{see2017get} and extended by \citet{gehrmann2018bottom}, versus the one by 
\citet{paulus2017deep}. In terms of ROUGE, we observe better results than those reported by \citet{paulus2017deep} -- see Table~\ref{tab:big_table} -- indicating a possible edge for the architecture proposed by \citet{see2017get}.

\bibliographystyle{acl_natbib}
\bibliography{emnlp-ijcnlp-2019}

\end{document}